# Structural Dilemmas and Developmental Pathways of Legal Argument Mining in the Era of Artificial Intelligence

Xianglei Liao[1], Chuanyi Li[2], Kun Chen[1†]
1 Law School, Nanjing University, Nanjing 210093, China
2 Software Institute, Nanjing University, Nanjing 210093, China

## Abstract

Against the backdrop of rapid advances in artificial intelligence, legal argument mining has emerged as an important research area linking legal texts with intelligent analysis, carrying significant theoretical and practical implications. Existing studies have primarily developed along three dimensions: data, technology, and theory. At the data level, raw legal texts and annotated corpora constitute the foundational resources. At the technological level, research paradigms have evolved from rule-based systems and traditional machine learning to large language models (LLMs). At the theoretical level, argumentation theory and legal dogmatics provide important references for modeling argumentation structures. However, despite ongoing progress, the overall development of legal argument mining remains relatively slow. Building on a systematic review of existing research, this study conducts an in-depth analysis and finds that this is due not only to data scarcity or technical limitations, but more fundamentally to the lack of a structured representational approach that reconciles theoretical expressiveness with computational feasibility. Specifically, this challenge manifests in dilemmas in data standardization, obstacles to effective modeling, and limitations in domain adaptation. In response, the study proposes several key directions for future research. It aims to provide a reframing of key problems and a pathway for future development in legal argument mining, while leaving specific models and implementation schemes for further investigation.

† Corresponding Author: Kun Chen, Email: chenkun@nju.edu.cn.

## I. Introduction

Text mining and information extraction using artificial intelligence technologies have become a significant direction in contemporary interdisciplinary research and practice. Compared with general language texts, legal texts are typically characterized by a public nature, authority, and a high degree of normativity. In addition, they have significant value and exert substantial influence in social governance. The effective mining and utilization of such legal data are not only crucial to the development of intelligent judicial technologies, but also directly affect the quality and efficiency of the functioning of the rule of law. It is therefore of great importance for enhancing judicial transparency, standardizing judicial reasoning, and promoting the modernization of the rule of law.

At present, research on legal data mining can be broadly divided into two approaches. The first is a data-driven, knowledge inference-oriented approach, which relies on large-scale statistical learning and inductive pattern learning. Through techniques such as machine learning, deep learning, and large language models (LLMs), it extracts regularities and produces predictive results, such as sentencing prediction [1] and judicial decision prediction [2]. The second is a light theory-guided data mining approach, which draws on legal theory to provide a framework for data mining and result analysis [3]. This approach focuses on legal texts and argumentation models and identifies key informational elements such as argumentative structures, rhetorical structures, and points of dispute. Representative tasks include legal text summarization [4], argument sentence identification [5], legal interpretation classification [6], and legal argument structure mining [7]. Legal argument mining refers to the process of automatically identifying and structurally representing argumentative information in legal texts with computational techniques. Legal practice is essentially an argumentative activity. This is particularly evident in judicial decisions and legal opinions, where argumentative information forms the core of legal reasoning and judicial justification. Effective extraction of such information is therefore crucial for the full exploitation of legal data. Moreover, with the development of automated reasoning systems and legal education support tools, argument mining has also become a crucial prerequisite for the construction of such systems.

Since the late twentieth century, research on legal argument mining has gradually attracted scholarly attention [4,8]. Building on the seminal work of Moens and Palau, who proposed three core tasks of legal argument mining, the field has developed into an important topic at the intersection of artificial intelligence and law [7]. Over the past three decades, significant progress has been made in terms of data accumulation, technological approaches, and theoretical integration. Nevertheless, overall progress in practice has remained relatively slow. This paper identifies several underlying reasons for this situation. First, the lack of unified annotation standards makes it difficult to standardize data and effectively utilize datasets. Second, limitations in current computational capabilities hinder the modeling of complex argument structures. Third, domain-specific challenges impede accuracy, rigor, interpretability, and reliable evaluation. Finally, and more fundamentally, there exists a structural deficiency, namely the absence of a structured representation that bridges the expressive requirements of legal theory with computational feasibility. This absence prevents effective integration among data, models, and technologies.

Against this background, this study offers a systematic review of existing research and, through theoretical reflection, identifies and analyzes the structural bottlenecks in legal argument mining. It then proposes several directions for future research. The aim is to establish a clear problem framework and a coherent research agenda for the field's further development.

## II. Research Progress in Legal Argument Mining

The core issue of legal argumentation mining is how to effectively identify or extract information of argumentative significance from legal texts using intelligent technologies. As research has advanced, approaches that rely solely on data and technical methods have become increasingly inadequate to address the specialization and complexity of legal argument. As a result, some studies have begun to draw on argumentation theory or legal doctrine as complementary frameworks. Across the literature, legal argument mining is driven jointly by data, technology, and theory. The following discussion examines research developments in this field from these three dimensions.

### 1. Data: From Raw Texts to Annotated Corpora

The quantity and quality of data directly affect the precision and granularity of legal argument mining. From the perspective of existing research paradigms, relevant data can be broadly divided into raw data and annotated data, with the key distinction being whether explicit information about argumentation structures is included.

Raw data consists of legal texts without semantic annotation, typically in unstructured or semi-structured formats. Due to the high demands for informational accuracy in the legal domain, the sources of such data are relatively concentrated, mainly including officially published judicial decisions, statutes and regulations, court reports, parliamentary records, and certain legal commentaries. Among these, judicial decisions and legal opinions, given their relatively complete argumentative structures, constitute the primary objects of study. In practice, influenced by factors such as text length and the distribution of argumentative structures, researchers sometimes focus only on specific sections of judicial decisions, such as the section Findings of the Court [9], the Law section [10], or case summaries [11]. In addition, from the perspective of language distribution, existing research remains predominantly focused on English-language legal texts; although other languages are gradually being explored, their overall scale and level of maturity remain relatively limited.

Annotated data constitutes the essential resource for legal argument mining. Such data are typically produced through manual annotation of raw texts according to predefined guidelines, or further expanded by incorporating model outputs on this basis. The annotation process itself forms a crucial preliminary stage of argument mining. Its quality directly affects subsequent model training and evaluation. Depending on the content of annotation, existing corpora can be broadly categorized into three types: (1) annotations targeting specific types of argument, such as particular legal argument schemes [12]; (2) framework-oriented annotations centered on the argumentative process, aiming to represent participants, argumentative content, and outcomes [13]; and (3) annotations of argumentation structure, focusing on premises, conclusions, and their interrelations [14]. As research in this field has progressed, the granularity of annotation has evolved from a coarse-grained distinction between argumentative and non-argumentative sentences to a basic distinction between premises and conclusions, and ultimately to multi-component structural annotation schemes incorporating relations such as support and attack [15].

Despite the significantly higher value of annotated corpora compared to raw data, they are costly to produce and limited in scale. Inconsistencies in annotation standards across datasets are common. Moreover, variations in language and legal systems further constrain the cross-domain applicability of such corpora. In general, the current landscape at the data level reveals a clear imbalance. Legal argument mining is characterized by an abundance of raw data with insufficient structural information, alongside annotated data that is highly valuable but limited in scale.

**2. Technology: From Rule-Based Methods to LLMs**

From a technical perspective, the evolution of legal argument mining can be broadly divided into three stages: an early rule-based phase, a middle stage driven by traditional machine learning methods grounded in feature engineering, and a more recent shift toward deep learning–based approaches with LLMs at their core. It should be noted that this evolution does not represent a simple replacement of earlier methods by later ones, but rather a shift in research focus and technological mainstream. Different methods continue to have their respective domains of applicability across tasks, languages, and data conditions.

In its early phase, legal argument mining relied primarily on rule-based design and knowledge engineering approaches, as exemplified by early automatic legal text summarization systems such as SALOMON [4] for Belgian criminal cases. This stage was marked by strong interpretability and the ability to reflect explicit legal knowledge structures in specific tasks. However, its limitations were also evident. Rule design depended heavily on human expertise and domain knowledge, resulting in weak transferability and scalability, and making it difficult to adapt to the complex, diverse, and evolving forms of expression found in legal texts.

With the development of traditional machine learning, legal argument mining gradually entered a stage centered on manual feature design and classifier training. For instance, a series of studies by Moens and her colleagues employed methods such as Naive Bayes (NB), Maximum Entropy (MaxEnt), and Support Vector Machines (SVM) to perform different tasks, including identifying argumentative sentences, classifying premises and conclusions, and recognizing argumentation structures in legal texts [5,7,10]. This stage marked the formal emergence of legal argument mining as a relatively independent research field. Compared with rule-based approaches, traditional machine learning methods reduce reliance on explicitly designed rules to some extent and improve model generalization capability. However, traditional machine learning methods rely heavily on manual feature engineering and are constrained by the limited representational capacity of shallow models in semantic understanding, contextual modeling, and implicit knowledge processing. As a result, they often struggle to accurately capture ellipsis, normative language, and complex reasoning structures that are common in legal texts. Accordingly, related research has increasingly shifted toward representation learning and deep learning-based approaches [16].

In recent years, breakthroughs in deep learning have significantly enhanced the capabilities of natural language processing in semantic representation and contextual modeling. This, in turn, has driven legal argument mining into a stage centered on deep neural networks. A key advantage of this approach lies in its reduced reliance on manually engineered features, instead enabling models to learn richer lexical, syntactic, and semantic representations from large-scale corpora. For example, in studies on argument component classification using the Canadian CanLII corpus, deep neural network models have been shown to outperform traditional models such as random forests [11]. Among deep learning approaches, the application of LLMs is of particular significance for legal argument mining. This is reflected not only in their strong zero-shot and few-shot learning capabilities, but also in their broader impact on the research paradigm of argument mining. LLMs have fundamentally reshaped the field of argument mining, shifting it from a pipeline of supervised, task-specific classifiers to more diverse paradigms characterized by prompt-based methods, retrieval-augmented approaches, and reasoning-oriented frameworks [17]. However, important limitations remain. Existing general-purpose LLMs are primarily based on large-scale representation learning from textual corpora, enabling them to capture complex linguistic patterns and contextual relationships; they exhibit relatively weak capacities for conceptual understanding and normative reasoning, and cannot genuinely comprehend legal theories, but only simulate their expression at the linguistic level [18]. In addition, such models face challenges in interpretability and robustness. Against this backdrop, the development of domain-specific pre-trained models in the legal field has further enhanced the adaptability of technology to legal language. Through continued pre-training or training from scratch on legal corpora, researchers have developed domain-specific models better suited to legal texts, among which Legal-BERT is a representative example [19]. Compared with general-purpose LLMs, these domain-pretrained models demonstrate stronger performance in handling legal terminology, normative expressions, and recurring structural patterns in legal texts. Studies suggest that domain pre-training is particularly effective for legal tasks characterized by limited data, high complexity, and strong domain specificity [20]. Empirical research further supports this view: domain-pretrained BERT variants combined with neural networks generally outperform other embedding methods such as GloVe and ELMo in argument mining tasks, and integrating neural networks with domain-pretrained BERT models can further improve classification performance [21]. The studies by Huihui Xu et al. integrate neural networks with domain-specific pre-trained BERT models, further improving classification performance [22,23].

Despite the growing prominence of deep neural network methods, traditional machine learning has not been entirely displaced in legal argument mining. From one perspective, in contexts where linguistic resources are relatively scarce or task scales are small, traditional methods retain advantages in terms of stability, controllability, and lower cost, as evidenced by practices in Japanese legal argument mining [14]. From another perspective, traditional classifiers are often incorporated into hybrid approaches alongside neural networks or pre-trained models. For example, some studies adopt a combined technical framework integrating neural networks, pre-trained models, and support vector machines [24]. This suggests that, in the highly specialized task of legal argument mining, technological evolution is better understood as the optimization of multiple methods in combination, rather than the complete replacement of one paradigm by another.

On the whole, the technical foundations of legal argument mining have evolved alongside advances in artificial intelligence, with deep neural networks—particularly LLMs—emerging as the dominant paradigm. However, improvements in technical performance do not equate to a full resolution of the challenges posed by legal argumentation structures. Due to the inherent complexity of such structures, existing methods still exhibit clear limitations. Empirical studies show that even LLMs enhanced through fine-tuning or prompting strategies perform worse than RoBERTa baselines (F1=41.3%) in identifying fine-grained argument types in legal texts [25]. Although domain-pretrained LLMs achieve higher accuracy, they also entail substantial computational costs. In general, while current technologies achieve relatively high accuracy in identifying premises and conclusions, they remain less effective in detecting elements such as facts, norms, exceptions, and specific premises associated with argument schemes, as well as support and attack relations, and nested structures composed of multiple propositions and relations (e.g., propositions A and B jointly support proposition D, mediated by their relations to proposition C). This indicates that rapid advances in technology have not automatically translated into corresponding improvements in the representation of argumentation structures, and that a unified and stable foundation for structural representation across different technical approaches is still lacking. As a result, although technological development in legal argument mining continues to advance, further breakthroughs will depend on closer integration with structural modeling and data standardization.

## 3. Theory: From Data-Driven Approaches to the Integration of Domain Theory

In contrast to the rapid development at the data and technical levels, systematic theoretical construction has long remained relatively weak in research on legal argument mining. Previous studies have largely focused on improving model performance and implementing specific tasks, while paying insufficient attention to the underlying theories of argumentation on which such mining depends. Typically, these studies operate on existing legal texts, extracting argumentation structures or performing pattern recognition without relying on an explicit argumentation framework [13,26]. Although such approaches are valuable for summarizing types of arguments and describing practical features, they remain limited in important respects. This is mainly due to their lack of domain-theoretical norms and conceptual foundations. As a result, they face clear limitations in deep structural representation, normative evaluation, and the modeling of complex reasoning.

Some studies have recognized this issue and have incorporated argumentation theory or legal dogmatics to address the limitations of purely data-driven approaches, thereby giving rise to a research orientation characterized by the coordinated integration of domain theory and data-driven methods. In terms of general argumentation theory, Rhetorical Structure Theory (RST) has been applied to analyze semantic and functional relations within texts. Its core lies in characterizing textual organization and coherence through rhetorical relations, while emphasizing the connection between text structure and persuasive effect. As early as 2007, Moens and colleagues conducted studies on the identification of argumentative and non-argumentative sentences in legal texts based on this theoretical framework [5]. Pragma-dialectics, by contrast, approaches argumentation from the perspective of critical discussion, conceiving it as an interactive process centered on a specific standpoint. It classifies arguments into types, including single, multiple, coordinative, and subordinative, depending on how premises support conclusions [27]. In addition, research grounded in legal dogmatics focuses on extracting specific normative structures in legal texts, such as forms of legal interpretation [6] and the application of principles like proportionality [28].

With regard to formal models of argumentation, the Toulmin model, argumentation schemes, and the Carneades model have been widely used for the structured representation of legal argumentation. The Toulmin model refines the internal structure of arguments by distinguishing elements such as data, warrant, backing, modal qualifiers, and rebuttals [29]. In practice, however, due to constraints related to model complexity and mining capabilities, studies often adopt simplified or modified versions of this model. Argumentation scheme theory provides a typological approach centered on argument templates, which not only capture structural features but also introduce critical questions to assess the applicability and soundness of arguments [30]. Since there is diversity and relative independence among different schemes, this theory can be used to identify specific types of arguments, extract structural elements, or combine both functions [31]. The Carneades model further refines argumentation structure by distinguishing among ordinary premises, assumptions, and exceptions, and by differentiating between supporting and opposing arguments [32]. It is currently one of the more expressive and complex formal models. Argumentation support systems built on this model, namely the Carneades System, have been applied in related research [33]. In recent years, scholars have also attempted to incorporate this model into modeling practices in legal argument mining.

What the current literature indicates is that the introduction of domain theory has, to some extent, improved the representation of argument structures in legal argument mining, enabling it to move beyond surface-level pattern recognition. However, differences among theoretical models in terms of conceptual distinctions, structural granularity, and modes of representation make it difficult to translate them directly into a unified computational framework. Moreover, there is considerable variation across studies in the selection and simplification of theoretical models, and effective coordination between theoretical modeling for argumentation mining and argument representation has yet to be achieved. This pluralistic yet fragmented situation means that, although theory has been widely introduced, it has not yet formed a stable and generalizable foundation for structural representation, thereby limiting its practical utility in data annotation and model construction.

## 4. Summary

Legal argument mining has made notable progress across the three dimensions of data, technology, and theory (see Table 1). At the data level, legal text resources have become increasingly abundant, and annotated corpora have gradually accumulated. At the technical level, research has evolved from rule-based methods to traditional machine learning and further to deep neural networks and LLMs, with continuous improvements in model performance. At the theoretical level, argumentation theory and legal dogmatics have been progressively introduced, promoting a shift from data-driven approaches to those jointly driven by domain theory and data.

However, developments across these three dimensions remain, on the whole, somewhat fragmented. At the data level, annotation standards have yet to be unified, making it difficult to achieve cross-corpus sharing and reuse. At the technical level, although model capabilities continue to improve, limitations persist in the representation of argument structures and the cumulative usability of results. At the theoretical level, the pluralistic introduction of frameworks lacks a unified mode of structural representation and an integrative pathway. It is thus evident that, despite progress across multiple dimensions, stable mechanisms for coordination among data, technology, and theory have yet to be established, which in turn constrains the further development of legal argumentation mining.

Table 1: Comparative Study of Legal Argument Mining

| Raw Data | Annotation Targets | Annotation Content | Task Types | Technical Approaches | Theoretical Foundations | Representative Literature |
|---|---|---|---|---|---|---|
| Belgian criminal cases | Case metadata | The name of the court, The decision date, The offences charged, Statutory provisions, Legal principles | Information extraction | Text grammar | None | Moens & Uyttendaele [4], Uyttendaele [34] |
| Araucaria corpus | Components | Argumentative sentences, Non-argumentative sentences | Classification | Multinomial naive Bayes classifier, Maximum entropy model | Argumentation theory | Moens et al. [5] |
| Legal documents from ECHR | Components | Premises, Conclusions | Classification | Context-free grammar | Argumentation theory | Palau & Moens [35] |
| Araucaria corpus, ECHR corpus | Component, Relations | Premises, Conclusions, Argumentation structures | Classification, Structure identification | Naive Bayes classifier, Maximum entropy model, Support vector machines, Context-free grammar | Argumentation theory | Palau & Moens [7] |
| Japanese civil law judgement documents | Component, Relations | Issue Topic Identification, Rhetorical Classification, Issue Topic Linking, FRAMING Linking | Classification, Relation identification | Support vector machines, Conditional random fields | Argumentation theory | Yamada et al. [14] |
| Cases and summaries from CanLII | Components | Issues, Reasons, Conclusion, Non-IRC. | Classification | Traditional Machine Learning (random forest), Deep Neural Networks, FastText | None | Xu et al. [11] |
| Chinese judgement documents | Dialogical structure | Interactive Argument-Pair | Information extraction | Pretrained language model, Fine-tuning mechanisms | None | Yuan et al. [13] |
| Decisions of CJEU | Component, Schemes | Premises, Conclusions, The type of premise (legal or factual), Argument schemes | Classification, Schemes identification | Linear svc, SVC, Random forest, Gaussian naive Bayes, K-neighbours, TF-IDF, SBERT, Legal-BERT | Argumentation theory | Grundler et al. [9] |
| ECHR decisions | Case metadata, Schemes | Actors, Argument types | Information extraction, Schemes identification | RoBERTa-Large, Legal-BERT | Argumentation theory | Habernal et al. [24] |
| GFCC decisions | Normative structure | Elements of proportionality principle | Structure identification | BERT, LSTM, CRF | Legal dogmatics (proportionality principle) | Lüders & Stohlmann [28] |
| Italian decisions on Value Added Tax | Component, Schemes | Premises, Conclusions, The type of premise (legal vs factual), Argument schemes | Classification, Schemes identification | LLM | Argumentation theory | Grundler et al. [31] |

| Raw Data | Annotation Targets | Annotation Content | Task Types | Technical Approaches | Theoretical Foundations | Representative Literature |
|---|---|---|---|---|---|---|
| Indian / UK case documents | Components | Facts, Ruling by lower court, Argument, Ratio of the decision, Statute, Precedents, Ruling by the present court | Information extraction | MARRO models | None | Bambroo et al. [26] |
| Indian Supreme Court judgements | Relations | Support relations, Attack relations | Relation identification | Rule-based methods, BERT | None | Ali et al. [36] |
| ECtHR judgments | Interpretive techniques | Types of legal interpretation | Classification | LLMs | Legal dogmatics (theory of legal interpretation) | Dugac & Altwicker [6] |

## III. Core Challenges in Legal Argument Mining

As noted above, existing research remains highly fragmented in terms of data sources, annotation modes, structural representations, and theoretical foundations, with a lack of unified representational bases and integration mechanisms across different lines of research. Further analysis suggests that research on legal argument mining is not determined by any single technical or resource bottleneck alone. Rather, it is shaped by a set of interrelated structural constraints, which are mainly reflected in three aspects: data and annotation standards; structural representation and computational modeling; and domain adaptation, evaluation, and validation.

### 1. Challenges in Data and Annotation Standards

Legal argument mining first faces fundamental constraints at the level of data and annotation. As previously discussed, the current state of data in argument mining can be characterized as "abundant raw data but insufficient structural information, and high-value annotated data but limited in scale." For fine-grained argument mining tasks that require the identification of argumentative propositions and their relations, high-quality annotated corpora are, in fact, an indispensable prerequisite.

However, legal texts are inherently long, complex in expression, and highly implicit in meaning. They are also often characterized by non-standard expressions, a mixture of argumentative and narrative discourse, and the omission of key information. In addition, both legal knowledge and argumentation knowledge are highly specialized, making the manual annotation of argumentative information extremely costly. Furthermore, annotation of similar cases often involves interpretive disagreements, making it difficult to establish consistent standards and thereby increasing the overall complexity of annotation work. To reduce the cost of manual annotation, some studies have sought to leverage LLMs for legal argumentation annotation, followed by human verification [37]. Although LLM-based annotation can substantially improve efficiency, the design of annotation guidelines remains a prerequisite for legal argument annotation. These guidelines define the specific tasks, requirements, and standards of annotation. Moreover, their development, validation, and refinement rely on feedback derived from manually annotated data. This implies that manually annotated data remains a necessary prerequisite for LLM-based mining. Moreover, the challenges of textual complexity and annotation disagreement that affect human annotators also extend to LLMs, resulting in significantly lower quality of LLM-generated annotations in fine-grained argument annotation tasks compared with those produced by experts. Under conditions of low-quality automatic annotation, a pipeline based on "automatic annotation followed by human verification" may not outperform a more incremental approach, such as "small-scale manual annotation, followed by machine learning, then automatic annotation, and subsequent human review." The fundamental reason lies in the fact that the elements of argument mining are explicitly defined. Statistical and computational approaches in machine learning are better suited to learning from cases by identifying and generalizing patterns of similarity, rather than interpreting and applying abstract guideline-based instructions for decision-making. Accordingly, in highly complex argument mining tasks, manually annotated data remains indispensable. However, due to the inherent difficulties of human annotation and the insufficient capability of large models to produce high-quality automatic annotations, genuinely high-quality annotated corpora suitable for training and evaluation remain scarce. This problem has been identified by some scholars as a fundamental obstacle to legal argument mining [38].

Beyond the difficulty of producing annotated data, a more critical issue lies in the lack of unified standards. On the one hand, publicly available annotation guidelines are relatively scarce, which leads to annotation schemes being heavily dependent on specific research contexts. On the other hand, variations across corpora in terms of annotation units, granularity, and representational structures make it difficult to integrate and reuse data effectively. For instance, some studies adopt sentences as the basic unit, others use argumentative components or argument types, while still others directly annotate argumentative relations or discourse structures. Such heterogeneity hinders alignment and integration across datasets, thereby limiting data sharing, reuse, and the cumulative development and comparability of research findings. In this sense, the "insufficiency of data" is merely a surface manifestation of the problem; the more fundamental issue lies in the lack of standardized annotation schemes.

**2. Challenges in Structural Representation and Computational Modeling**

In addition to data constraints, a core difficulty in legal argumentation mining lies in the tension between structural representation and computational modeling. On the one hand, legal argumentation theory has developed a variety of fine-grained structural models, such as the Toulmin model, argumentation schemes, and the Carneades model, which can comprehensively capture both the structure of argumentation and variations in argument types. However, argumentative expression in judicial decisions does not strictly conform to these theoretical models and instead exhibits a high degree of flexibility and hybridity. For example, judges may sometimes derive conclusions directly from norms and facts, reflecting a clear syllogistic structure of judicial reasoning. At other times, however, they may engage in interpretive reasoning regarding legal norms, develop arguments in the process of fact-finding, or respond to the claims of the parties. At the technical level, in order to achieve computational tractability, available models often simplify argumentation structures, reducing them to binary "premise–conclusion" forms, ternary "issue–reason–conclusion" structures, or "facts–law–conclusion" representations. While such simplifications facilitate model training and improve performance, they also weaken the ability to represent the complexity of argumentative structures found in legal practice. This gives rise to a typical tension: legal argumentation theory foregrounds structural richness and granularity, whereas computational models generally privilege low-dimensional, homogeneous representations for tractability. This tension has also been described as one in which computational modeling reduces argumentation to general premises and claims, while legal scholarship characterizes argumentation as exhibiting rich typological diversity.

Furthermore, at a more granular level, there is also a mismatch between different layers of argument mining. Annotation in datasets is often conducted at the level of sentences or components; theoretical models distinguish multiple argumentative elements and relations; and technical implementations tend to frame the problem as coarser-grained classification tasks. This misalignment in granularity across data, theory, and models makes it difficult to represent argumentative structures in a stable and complete manner. Consequently, how to construct an intermediate representation that both captures the complexity of legal argumentation and remains computationally tractable has become a central issue in legal argumentation mining.

## 3. Challenges in Domain Adaptation and Evaluation

Argument mining in the legal domain faces significant cross-domain adaptation challenges. As an interdisciplinary field situated at the intersection of law and artificial intelligence, legal argument mining operates within a normative practice in which legal texts are shaped by institutional constraints, often saturated with value judgments, and subject to heightened requirements of accuracy, rigor, and interpretability. However, existing technical approaches remain poorly adapted to these domain-specific demands.

First, at the level of cognitive and inferential paradigms, contemporary AI approaches are predominantly grounded in statistical correlation modeling, whereas legal argumentation is rooted in normative reasoning. The latter foregrounds features such as defeasibility, justificatory structure, and sensitivity to institutional constraints, thereby giving rise to a fundamental methodological divergence. This divergence has direct implications for how models identify and interpret argumentative relations.

Second, current LLMs continue to exhibit a non-negligible degree of hallucination and bias, which poses a significant obstacle to meeting the accuracy demands of the legal domain. Legal argumentation is inherently structured, systematic, and interdependent, such that an error in one component can propagate through the reasoning chain and compromise both the overall structure and the final conclusion. Compounding this concern, legal decision-making is a high-stakes activity with an exceptionally low tolerance for epistemic error. Inaccuracies in model outputs or extracted argumentative structures may directly affect determinations of rights and obligations, thereby undermining judicial legitimacy. Accordingly, legal argument mining entails far more stringent requirements with respect to output reliability.

Third, legal argumentation—especially in hard cases—involves extensive value judgments and interest balancing [39], which are often implicit in the text, such as appeals to common sense, public policy, or the principle of proportionality. Machine systems face considerable difficulty in identifying and interpreting such implicit elements in the absence of explicit linguistic markers. In addition, the recognition of argumentative structures depends on a holistic understanding of contextual information, including intra-textual relations as well as broader institutional and socio-legal backgrounds. As Marie-Francine Moens notes, humans rely on world knowledge, common sense, and domain-specific knowledge to identify argumentative discourse; a major obstacle to building high-performance argument mining systems is the lack of such knowledge [16]. Although LLMs have recently made significant advances in language processing capabilities, they still exhibit limitations in long-range context integration, reconstruction of implicit premises, and the interpretation of normative semantics.

Legal argument mining also faces challenges in the post-extraction evaluation and validation of results. At present, there is no unified evaluation standard or normative framework for assessing extracted legal arguments, and different modeling approaches produce substantially different outputs. In evaluation, variations in the selected criteria can significantly affect results; without standardized benchmarks, it becomes difficult to compare or reproduce findings across studies. This inconsistency in evaluation prevents improvements in model performance from being accumulated into verifiable and systematic research progress.

In addition, considerations of cost and systematization suggest that argument computation and argument evaluation should be automated and tightly integrated with argument mining models. At present, however, research exhibits a fragmented development pattern: argument computation, argument evaluation, and argument mining are largely pursued in isolation, and existing evaluation and computation frameworks remain poorly aligned with mining models.

**4. Structural Root Cause: The Absence of an Intermediate Representational Layer**

Taken together, the three dimensions discussed above—data, modeling, and domain adaptation—are not isolated problems. Rather, they converge on a deeper structural issue: the absence of a stable "representation layer" in legal argument mining, namely, a structured intermediate representation that can serve as a bridge between legal theoretical formulations, annotation practices, and computational models.

More specifically, at the data level, the lack of a unified representational scheme leads to substantial heterogeneity in annotation targets, granularity, and guidelines across different studies, making annotated corpora difficult to align, share, and reuse. At the modeling level, in the absence of a stable structural representation, computational models tend either to oversimplify argumentation into low-dimensional structures or fail to adequately accommodate complex theoretical frameworks of argumentation, thereby generating a tension between representational expressiveness and computational tractability. At the level of domain adaptation, the lack of formalization of argumentative structures as computable objects limits models' ability to systematically incorporate normative constraints, value judgments, and contextual information into their processing.

From a methodological perspective, this challenge can be understood as a missing shared representational layer linking theory, data, and models in legal argument mining: theoretical constructs are difficult to translate into operational annotation guidelines; annotated data are insufficient to support stable model training; and model outputs, in turn, are difficult to map back into theoretical frameworks for interpretation and evaluation. The absence of such a representation layer prevents different strands of research from generating cumulative progress and constrains the field from evolving from isolated tasks toward a more systematic and integrated research program.

## IV. Advancement Pathways for Legal Argument Mining: Toward Structured Intermediate Representations

Building on the above analysis, the key direction for future research in legal argument mining does not lie in merely improving model performance or scaling up data, but rather in developing a structured intermediate representation that can bridge legal theory, annotation practice, and computational modeling. On this basis, the research paradigm of legal argument mining should be reconfigured accordingly. Within this framework, legal argument mining should no longer be conceived as a collection of isolated tasks (e.g., argument detection or relation classification), but rather as a system-oriented endeavor centered on structural representation. Accordingly, data acquisition, annotation design, model development, and result interpretation should all be organized around a unified representational framework, thereby enabling coordination and integration across tasks.

### 1. Constructing a Scalable Framework for Legal Argument Structure Representation

To begin with, it is necessary to introduce an extensible structured representation framework between theory and computation, in order to provide a unified description of the basic elements of legal argumentation and their interrelations.

Unlike traditional representation schemes that operate at the sentence level and focus primarily on a simple "premise–conclusion" dichotomy, such a structured representation should take propositions as the fundamental unit and clearly distinguish different types of propositions. It should be capable of accommodating a richer set of argumentative relations, including support, attack, combination, match, and identity relations, and should further allow these relations to be nested, thereby progressively approximating the real structure of legal argumentation. At the same time, the framework should maintain strong computational tractability, so that it can be effectively used both for human annotation and for machine learning models. A core feature of this representational framework is its extensibility: as computational capacity increases, the structure should be able to incrementally incorporate additional elements, enabling progressively finer-grained representations of information. Under such an extensible representation, argument mining results can develop in an accumulative manner, with models, data, and computation evolving within a dynamic and incremental process. One basic design principle of this structural representation is to establish "premise," "conclusion," and "support" as foundational elements, while introducing a placeholder such as "other" as an intermediate mechanism for extension, allowing further refinement of categories such as premises, conclusions, and supporting relations, and enabling the introduction of additional relations such as attack.

By introducing such a unified structured representation, it becomes possible to establish a common representational foundation across different datasets and tasks, thereby providing the prerequisite for subsequent data integration and accumulation, model training, and cross-system result comparison.

**2. Promoting the Development of Standardized Annotated Corpora for Legal Argument**

At present, research on annotated corpora in legal argument mining has primarily focused on increasing annotation volume and corpus size, resulting in the development of multiple corpora of different types and sizes [40]. However, the annotation schemes underlying these corpora are not standardized, and in many cases, the annotation guidelines are not publicly available, leading to limited reproducibility and the emergence of "data silos." Against this backdrop, future work should further advance the construction of standardized annotated corpora. This requires the adoption of a unified set of annotation rules, thereby enabling corpus construction within a consistent framework, reducing fragmentation, and supporting cumulative and scalable development.

The key to a unified annotation scheme lies in the formulation of clear and explicit annotation guidelines [41]. These guidelines should provide precise specifications for issues such as the categorization of proposition types, the identification of relation types, and annotation granularity. In addition, mechanisms such as annotator training, pilot annotation, double-blind independent annotation, and conflict resolution through adjudication should be employed to ensure annotation consistency. At the same time, all instances of ambiguity encountered during annotation and their corresponding adjudication decisions should be systematically documented, thereby forming a continuously evolving and self-refining annotation guideline system.

Furthermore, the construction of annotated corpora should gradually shift from a "single-task-oriented" paradigm to a "structure-oriented" paradigm. That is, rather than serving a specific model or task in isolation, annotated corpora should be designed as reusable data infrastructure that supports the joint development of multiple tasks, such as argument detection, structure identification, and reasoning simulation. This shift would also enhance both the theoretical and practical value of the corpora themselves.

**3. Strengthening Domain Knowledge–Driven Computational Models**

The identification of information such as parties, issues in dispute, statutory provisions, and factual circumstances by machine systems can largely be achieved based on the actual content and structure of legal data itself. However, the identification of legal argumentative elements and argumentative structures requires a deeper understanding of the relations among different components. At present, LLMs do not inherently possess such capabilities. Therefore, legal argument mining should shift from generic, data-driven approaches toward a domain-informed paradigm that tightly integrates legal knowledge with data-driven methods, thereby strengthening the shaping power of legal theoretical knowledge—particularly theories of legal argumentation—on computational models. To enhance the capacity of computational models to incorporate theoretically grounded knowledge, previous studies have proposed four main approaches: fine-tuning LLMs for specific tasks, domain-specific pretraining, retrieval-augmented generation (RAG), and the integration of LLMs with structured knowledge bases [17,42].

In the context of legal argument mining, fine-tuning refers to the adaptation of general-purpose LLMs to specific argument mining tasks through supervised learning on limited manually annotated argumentative data. Domain-specific pretraining involves further training a general model on large-scale legal corpora, typically complemented by a smaller set of annotated data. Retrieval-augmented generation (RAG) enables the model to access external legal knowledge sources—such as statutory databases and case law repositories—thereby improving the accuracy of argument understanding and extraction. Finally, approaches that integrate LLMs with structured knowledge bases incorporate formally represented legal knowledge to support more precise identification and analysis of argumentative structures.

As noted above, domain-pretrained models tend to achieve relatively strong performance in argument mining; however, they incur high training costs and must continuously cope with the evolving nature of legal knowledge, making them less than optimal in practice. Models relying solely on task-specific fine-tuning are significantly constrained by the lack of legal-domain knowledge, resulting in limited performance. A more promising direction is to develop a hybrid approach that combines retrieval-augmented generation , structured knowledge base integration, and task-specific fine-tuning. Concretely, this involves fine-tuning a general-purpose LLM on a small amount of annotated argumentative data, while enabling it to access external resources during argument mining, including statutory databases, case law repositories, and structured legal knowledge bases.

Since most countries already maintain relatively well-developed statutory and case law databases, and small-scale annotated datasets can be constructed through manual annotation, this approach is feasible in practice. Accordingly, the core challenge lies in the development of structured knowledge bases. The construction of such knowledge bases primarily relies on legal knowledge graphs composed of entities and relations. The structured and logical properties of knowledge graphs can enhance the reasoning capabilities of LLMs in the legal domain [43], thereby facilitating fine-grained argument mining. In the process of argument mining, knowledge graphs serve two central functions: first, they provide domain knowledge that supports contextual understanding and reasoning over legal texts; Second, certain relatively stable categories of knowledge are embedded in the very structure of argument mining models. The construction of legal-domain knowledge graphs can follow both top-down and bottom-up approaches. The top-down approach begins with legal experts who identify and define key ontologies, including concepts, hierarchical relations, and relevant rules, followed by entity learning to incorporate instances into the predefined conceptual system. For example, legal experts may systematically organize legal argumentation theory and doctrinal legal knowledge, transforming structurally significant elements into computable knowledge representations, which can then be used to impose structural constraints on models and to validate outputs. Alternatively, structured annotated data can be used to construct legal argumentation knowledge graphs by linking propositions, relations, and argumentation schemes, thereby supporting complex reasoning and explanation. The bottom-up approach, by contrast, relies on machine learning methods to detect and extract relevant legal terminology from case texts and to gradually induce, abstract, and organize them into hierarchical conceptual structures. Although machines are limited in extracting fine-grained argumentative structures, they are relatively reliable in identifying more basic informational elements, making them useful for preliminary extraction tasks.

Such a knowledge-enhanced paradigm—combining structured knowledge with annotated data—can effectively compensate for the deficiencies of general-purpose large language models in terms of interpretability and normativity, thereby enabling model outputs to better conform to the requirements of legal reasoning.

## 4. Exploring a Collaborative Research Paradigm among Legal Experts, Computational Scientists, and Machines

Human–machine collaboration is a fundamental principle for the application of artificial intelligence in advancing the rule of law. However, how to allocate tasks between humans and machines has long remained a difficult problem. With the rapid development of artificial intelligence, researchers have become increasingly reliant on data- and technology-driven approaches, a tendency that is even more pronounced in interdisciplinary research at the intersection of AI and the humanities and social sciences. As a field at the intersection of law and artificial intelligence, legal argument mining is characterized by a level of complexity that makes it neither fully automatable nor feasible to be completed solely by researchers from a single discipline. It is therefore necessary to construct a multi-agent collaborative model for both research and application, namely, to explore a "legal expert – computer scientist – machine" collaborative framework.

Specifically, legal experts are primarily responsible for the theoretical construction of argumentation structures, the formulation of annotation guidelines, the development of standardized annotated corpora of legal argumentation, and the normative evaluation of automated mining results. Computer scientists are responsible for building annotation platforms, designing mining models, maintaining systems, and implementing automated evaluation mechanisms. Machines, in turn, play a central role in large-scale data processing, pattern recognition, and evaluation and validation tasks. Through a clear division of labor and well-defined coordination mechanisms, research efficiency can be improved while maintaining legal rigor.

In particular, through bidirectional feedback loops between legal scholars and computer scientists mediated by machine-generated outputs, it is possible to gradually form the missing intermediate layer of structured representation in current legal argument mining research. For example, interactive annotation and analysis tools can be developed to enable legal scholars to actively participate in data annotation and model feedback processes, thereby integrating the expertise of both legal and computational researchers and realizing an iterative, co-constructed human–machine optimization cycle.

## 5. From Task-Oriented Approaches to Systematic Research Agendas

From the perspective of the overall research paradigm, legal argument mining should shift away from a task-centric research mode toward a more systematic research method centered on structured representation.

Prior studies are largely organized around individual tasks, such as argument identification, proposition classification, and relation detection. As Lawrence and Reed note in their survey, traditional argument mining approaches typically adopt a task-specific supervised pipeline, in which component detection, relation classification, and argument quality assessment are modeled as separate problems, supported by carefully designed annotation schemes and moderately sized corpora [44]. These originally holistic tasks have been artificially decomposed into fragmented sub-tasks, resulting in a corresponding fragmentation of data, theoretical frameworks, and technical outcomes. Consequently, the data, theoretical insights, and technical outputs generated in different studies are often fragmented, making it difficult to integrate and accumulate findings into a coherent and systematic research paradigm. Some recent studies have begun to combine multiple subtasks in argument mining in order to more comprehensively capture argumentative structures in legal texts. For example, Gechuan Zhang et al. first perform argument sentence identification to distinguish argumentative from non-argumentative sentences, then conduct argument relation extraction, and finally perform argument component classification [38]. Compared with single-task approaches, such combined pipelines are more consistent with the procedural nature of argument mining and are more conducive to cumulative and systematic development.

The three core tasks of argument mining—argument detection, proposition identification, and relation detection—proposed by Moens and Palau in 2009 laid the foundation of the field [7]. However, subsequent research has expanded this framework by incorporating argument scheme extraction [45] and the simulation of legal reasoning [46]. As a result, future research should, on the basis of a unified structured representation, integrate tasks such as argument detection, proposition classification, relation identification, argument scheme recognition, and reasoning simulation into a continuous and coherent framework, thereby enabling data sharing and methodological reuse across tasks.

More fundamentally, it is necessary to further integrate argument mining, argument computation, and argument evaluation around structured representation, so as to form a tightly coupled and systematically coordinated research ecosystem.

## V. Conclusion and Future Work

Driven by rapid advances in artificial intelligence, legal argument mining has become an increasingly important research direction connecting legal theory, judicial practice, and intelligent technologies. This paper provides a systematic review of existing studies from the perspectives of data, technology, and theory. It shows that, although some progress has been achieved, the field as a whole remains characterized by structural fragmentation, inconsistent standards, and limited cumulative development. Further analysis indicates that the central challenges in legal argument mining do not stem solely from insufficient data volume or limited model capacity. Rather, they lie in the absence of a structured intermediary representation that can serve as a bridge between legal theoretical formulations, annotation practices in textual data, and computational modeling. The lack of such an intermediate layer prevents the standardization of data, undermines the stability of model construction, and hinders the effective integration of theoretical insights, thereby constraining the systematic development of the field as a whole. In light of this, this paper proposes that future research should promote a transition from dispersed exploration to a more systematized framework by focusing on the construction of structured representation schemes, the standardization of annotated corpora, domain knowledge-driven modeling approaches, and mechanisms for human–machine collaboration. Compared with research paradigms oriented toward isolated tasks or individual technical pathways, this representation-centered approach is more conducive to achieving integration and coordination among data, technology, and theory.

It should be noted that this research primarily aims to synthesize the current state of studies, identify core challenges, and outline general directions for future work, rather than to develop specific technical implementations or system designs. The proposed frameworks for structured representation, annotation schemes, and their empirical applications therefore remain to be further developed and empirically validated in future research. It is hoped that, with the advancement of interdisciplinary collaboration, legal argument mining will foster a more robust interaction between theoretical development and technological implementation, thereby providing strong support for the advancement of legal practice and intelligent judicial systems.

[1] Seena Fazel, et al.,The predictive performance of criminal risk assessment tools used atsentencing: Systematic review of validation studies, Journal of Criminal Justice, Vol.81,2022,101902.
[2] Masha Medvedeva, Michel Vols & Martijn Wieling , Using machine learning to predict decisions of the European Court of Human Rights. Artificial Intelligence and Law,Vol. 28, 2020, pp.237-266.
[3] Lusheng Wang, On the Construction of "Domain Theory" of Legal Big Data(in Chinese), China Legal Science,No.2, 2020, pp.268-269.
[4] Marie-Francine Moens & Caroline Uyttendaele, Automatic Text Structuring and Categorization as a First Step in Summarizing Legal Cases, Information Processing & Management, Vol.33, 1997, pp.727-737.
[5] Marie-Francine Moens, et al., Automatic detection of arguments in legal texts, in: Proceedings of the 11th International Conference on Artificial Intelligence and Law, ACM Press, 2007, pp. 225-230.
[6] Gaspar Dugac & Tilmann Altwicker, Classifying legal interpretations using large language models, Artificial Intelligence and Law,Vol. 33, 2025.
[7] Raquel Mochales Palau & Marie-Francine Moens, Argumentation Mining: The Detection, Classification and Structure of Arguments in Text, in Proceedings of the 11th International Conference on Artificial Intelligence and Law , ACM Press, 2009, pp. 98-107.
[8] Kathleen Freeman & Arthur M. Farley, A model of argumentation and its application to legal reasoning, Artificial Intelligence and Law, Vol.4, pp.163–197.
[9] Giulia Grundler, et al., Detecting arguments in CJEU decisions on fiscal state aid, in: Proceedings of the 9th Workshop on Argument Mining, 2022, pp.143-157.
[10] Raquel Mochales Palau & Marie-Francine Moens, Study on the structure of argumentation in case law, in: Proceedings of the 21st International Conference on Legal Knowledge and Information Systems, IOS Press, 2008, pp.11-20.
[11] Huihui Xu, Jaromír Šavelka & Kevin D. Ashley, Using argument mining for legal text summarization, in: Legal Knowledge and Information Systems, Vol. 334, IOS Press, 2020, pp.184-193.
[12] Douglas Walton, Argument mining by applying argumentation schemes, Studies in Logic,Vol. 4, 2012, pp.38-64.
[13] Jian Yuan, et al., Overview of SMP-CAIL2020-Argmine: The Interactive Argument-Pair Extraction in Judgement Document Challenge, Data Intelligence,Vol.3, 2021, pp.287-307.
[14] Hiroaki Yamada, Simone Teufel & Takenobu Tokunaga, Building a corpus of legal argumentation in Japanese judgement documents: towards structure-based summarisation, Artificial Intelligence and Law,Vol. 27, 2019, pp.141-170.
[15] R¯uta Liepina, et al., Legal argument mining: recent trends and open challenges, in: Proceedings of the First Argument Mining and Empirical Legal Research Workshop, 2025.
[16] Marie-Francine Moens, Argumentation mining: how can a machine acquire common sense and world knowledge?, Argument & Computation,Vol.9, 2018, pp.1-14.
[17] Hao Li, et al.,Large Language Models in Argument Mining: A Survey, arXiv:2506.16383 [cs.CL].
[18] Weikang Yuan, et al.,Can Large Language Models Grasp Legal Theories? : Enhance Legal Reasoning with Insights from Multi-Agent Collaboration,Findings of the Association for Computational Linguistics: EMNLP 2024, pp.7577-7597.
[19] Ilias Chalkidis, et al., LEGAL-BERT: the muppets straight out of law school, in: Findings of the Association for Computational Linguistics: EMNLP 2020, 2020, pp. 2898–2904.
[20] Lucia Zheng, et al., When does pretraining help? Assessing self-supervised learning for law and the CaseHOLD dataset of 53,000+ legal holdings, in: Proceedings of the Eighteenth International Conference on Artificial Intelligence and Law, 2021, pp. 159-168.
[21] Gechuan Zhang, Paul Nulty & David Lillis, Enhancing legal argument mining with domain pre-training and neural networks, Journal of Data Mining and Digital Humanities, 2022.
[22] Huihui Xu, Jaromir Savelka & Kevin D. Ashley, Accounting for sentence position and legal domain sentence embedding in learning to classify case sentences, in: Legal Knowledge and Information Systems, Vol. 346, IOS Press, 2021, pp. 33–42.
[23] Huihui Xu, Jaromir Savelka & Kevin D. Ashley, Toward summarizing case decisions via extracting argument issues, reasons, and conclusions, in: Proceedings of the Eighteenth International Conference on Artificial Intelligence and Law, ACM Press, 2021, pp. 250–254.
[24] Ivan Habernal, et al., Mining legal arguments in court decisions, Artificial Intelligence and Law, Vol.31, 2023, pp.557-594.
[25] Lena Hel & Ivan Habernal, Contemporary LLMs struggle with extracting formal legal arguments, in: Proceedings of the Natural Legal Language Processing Workshop 2025, 2025, pp.292-303.
[26] Purbid Bambroo, et al., MARRO: multi-headed attention for rhetorical role labeling in legal documents, arXiv:2503.10659v1 [cs.CL] 08 Mar 2025.
[27] Frans H. van Eemeren ,Rob Grootendorst & A. Francisca Snoeck Henkemans, Argumentation: analysis, evaluation, presentation, Lawrence Erlbaum Associates, 2002, pp. 64–66.
[28] Kilian Lüders & Bent Stohlmann, Classifying proportionality - identification of a legal argument, Artificial Intelligence and Law,Vol.33, 2025, pp.1051-1078.
[29] Stephen E. Toulmin, The uses of argument, Cambridge University Press, 2003, pp.87-95.
[30] Kevin D. Ashley, Artificial intelligence and legal analytics: new tools for law practice in the digital age,

Cambridge University Press, 2017, p. 130.
[31] Giulia Grundler, et al., AMELIA-Argument Mining Evaluation on Legal documents in ItAlian: A CALAMITA challenge, in: Proceedings of the 10th Italian Conference on Computational Linguistics, Pisa, Italy, 2024.
[32] Thomas F. Gordon, Henry Prakken & Douglas Walton, The Carneades model of argument and burden of proof, Artificial Intelligence,Vol. 171, 2007,pp. 875-881.
[33] Douglas Walton, Argument Evaluation and Evidence, Springer International Publishing, 2016, p.126-129.
[34] Catherine Uyttendaele, Marie-Francine Moens & Jos Dumortier, SALOMON: Automatic Abstracting of Legal Cases for Effective Access to Court Decisions, Artificial Intelligence and Law, 1998,Vol.6, pp.59-79.
[35] Raquel Mochales Palau & Marie-Francine Moens, Study on the structure of argumentation in case law, in: Proceedings of the 21st International Conference on Legal Knowledge and Information Systems, IOS Press, 2008, pp. 11–20.
[36] Basit Ali, et al.,Constructing A Dataset of Support and Attack Relations in Legal Arguments in Court Judgements using Linguistic Rules, in: Proceedings of the 13th Conference on Language Resources and Evaluation (LREC 2022), pp.491-500.
[37] Serene Wang, Lavanya Pobbathi & Haihua Chen, LAMUS: A Large-Scale Corpus for Legal Argument Mining from U.S. Caselaw using LLMs, arXiv:2603.08286 [cs.CL].
[38] Gechuan Zhang, David Lillis & Paul Nulty, Can domain pre-training help interdisciplinary researchers from data annotation poverty? A case study of legal argument mining with bert-based transformers, in: Proceedings of the Workshop on Natural Language Processing for Digital Humanities, 2021, pp. 121-130.
[39] Eveline T. Feteris,Weighing and Balancing in the Justification of Judicial Decisions, Informal Logic, Vol.28, pp.20-30(2008).
[40] Gechuan Zhang, Paul Nulty & David Lillis, A Decade of Legal Argumentation Mining: Datasets and Approaches, in Paolo Rosso et al. (Eds.): Natural Language Processing and Information Systems, Springer, 2022, pp.240-252.
[41] Kun Chen, et al., Guidelines for the annotation and visualization of legal argumentation structures in Chinese judicial decisions, arXiv:2603.05171.
[42] Jérémie Cabessa, Hugo Hernault & Umer Mushtaq, Argument Mining with Fine-Tuned Large Language Models,in: Proceedings of the 31st International Conference on Computational Linguistics,2025, pp.6624-6635.
[43] Dezhao Song, et al., Knowledge Graph-Assisted LLM Post-Training for Enhanced Legal Reasoning, arXiv:2601.13806 [cs.CL].
[44] John Lawrence & Chris Reed, Argument Mining: A Survey, Computational Linguistics,Vol.45, pp.765-818.
[45] Vanessa Wei Feng & Graeme Hirst, Classifying arguments by scheme, in: Proceedings of the 49th Annual Meeting of the Association for Computational Linguistics, 2011, pp.987-996.
[46] Aleksander Smywiński-Pohl & Tomer Libal, Enhancing legal argument retrieval with optimized language model techniques, in: JSAI International Symposium on Artificial Intelligence, Springer, 2024, pp.93-108.